\documentclass[letterpaper]{article} 
\usepackage{aaai23}  
\usepackage{times}  
\usepackage{helvet}  
\usepackage{courier}  
\usepackage[hyphens]{url}  
\usepackage{graphicx} 
\usepackage{natbib}  
\usepackage{caption} 
\usepackage{algorithm}
\usepackage{algorithmic}
\usepackage{graphicx}
\usepackage{amsmath}
\usepackage{amssymb}
\usepackage{booktabs}
\usepackage{color}
\usepackage{multirow}
\usepackage{xcolor}
\usepackage{newfloat}
\usepackage{listings}
\DeclareCaptionStyle{ruled}{labelfont=normalfont,labelsep=colon,strut=off} 
\floatstyle{ruled}
\newfloat{listing}{tb}{lst}{}
\floatname{listing}{Listing}
\nocopyright 

\title{Frequency Regularization for Improving Adversarial Robustness}
\author {
    Binxiao Huang,
    Chaofan Tao,
    Rui Lin,
    Ngai Wong
}
\affiliations {
    Department of Electrical and Electronic Engineering, The University of Hong Kong\\~\\
    \{bxhuang, cftao, linrui, nwong\}@eee.hku.hk
}

\begin{document}

\maketitle

\begin{abstract}
Deep neural networks are incredibly vulnerable to crafted, human-imperceptible adversarial perturbations. Although adversarial training (AT) has proven to be an effective defense approach, we find that the AT-trained models heavily rely on the input low-frequency content for judgment, accounting for the low standard accuracy. To close the large gap between the standard and robust accuracies during AT, we investigate the frequency difference between clean and adversarial inputs, and propose a frequency regularization (FR) to align the output difference in the spectral domain. Besides, we find Stochastic Weight Averaging (SWA)~\cite{izmailov2018averaging}, by smoothing the kernels over epochs\footnote{The final weight for evaluation is the average of the weights of multiple checkpoints during the training process.}, further improves the robustness. Among various defense schemes, our method achieves the strongest robustness against attacks by PGD-20, C\&W and Autoattack, on a WideResNet trained on CIFAR-10 without any extra data. 
\end{abstract}

\section{Introduction}
\label{sec:intro}
Deep neural networks (DNNs) have exhibited strong capabilities in various  applications~\cite{he2016deep, bin2019mr, tao2020dynamic}. However, research in adversarial learning shows that even well-trained DNNs are highly vulnerable to carefully crafted, human-imperceptible perturbations~\cite{goodfellow2014explaining,szegedy2013intriguing}. 
Recently, various defense methods~\cite{zhang2019theoretically,wang2019improving,wu2020adversarial} have been proposed to improve the robustness. Adversarial training (AT)~\cite{madry2017towards}, as a min-max saddle point problem, proves to be an effective and promising defense method without obfuscated gradients problems~\cite{athalye2018obfuscated}. 
In the following, we denote the models obtained by natural training and AT as natural and robust models, respectively. For robust models, the accuracy achieved on natural and adversarial inputs are denoted as standard accuracy and robust accuracy, respectively.
While AT improves robust accuracies, it generally sacrifices standard accuracies. Besides, frequency analysis~\cite{wang2020high} has been explored to yield new insights into DNNs. \textit{In this work, we aim to answer the following questions using a frequency lens: 1) Why does AT reduce standard accuracy? and 2) how to improve the robustness by narrowing the gap between the standard and robust accuracies?} 

To this end, we apply low-pass filtering (LPF) to the natural and adversarial inputs. Empirical results demonstrate that the robust model mainly relies on low-frequency content for prediction, which accounts for the low standard accuracy as high-frequency information is ignored. We also discover that the white-box attack can adapt its aggressive frequency distribution to the target model's frequency bias, thus explaining why white-box attacks are hard to defend.
By visualizing the differences between the natural and adversarial inputs, we reveal that the differences are mainly concentrated in the low-frequency region. In order to close the accuracy gap, we propose a frequency regularization (FR) that aligns the outputs for natural and adversarial inputs in the frequency domain, leading to improvement in the robust accuracy. In addition, by observing that the robust model has a smoother kernel than its natural counterpart, we employ Stochastic Weight Averaging (SWA)~\cite{izmailov2018averaging} as a method of smoothing kernels over the training steps to further improve robustness. 

To summarize, our work novelly adopts a frequency lens to: 1) explain the low standard accuracy of the robust model, and 2) propose a frequency-based regularization to significantly improve the robust accuracy.

\section{Related works}
\paragraph{Adversarial Defense.} Among various defense methods that have been proposed to improve robustness~\cite{szegedy2013intriguing,madry2017towards},
AT~\cite{athalye2018obfuscated} constitutes an effective and promising means. Typically, AT feeds adversarial inputs into a DNN to solve the following min-max optimization problem:
 
\begin{equation}
    \min _{\theta} \frac{1}{n} \sum_{i=1}^{n} \max _{\left\|\mathbf{x}_{i}^{\prime}-\mathbf{x}_{i}\right\|_{p} \leq \epsilon} \mathcal{L}\left(f_{\theta}\left(\mathbf{x}_{i}^{\prime}\right), y_{i}\right),
\end{equation}
where $n$ is the number of training examples, $\mathbf{x}_{i}^{\prime}$ is the adversarial input within the $\epsilon$-ball (bounded by an $L_p$-norm) centered at the natural input $\mathbf{x}_{i}$, $f_{\theta}$ is the DNN with weight $\theta$, $\mathcal{L}(\cdot)$ is the classification loss, \textcolor{black}{e.g., cross-entropy (CE).} Some recent results inspired by AT are also in place to further raise the robust accuracy: Zhang \emph{et al.}~\cite{zhang2019theoretically} \textcolor{
black}{identify} a \textcolor{
black}{tradeoff} between standard and robust accuracies that serves as a guiding principle for designing the defenses. 
Wu \emph{et al.}~\cite{wu2020adversarial} \textcolor{black}{identify} that the weight loss landscape is closely related to the robust generalization gap and \textcolor{black}{propose} an effective Adversarial Weight Perturbation (AWP) method to overcome the robust overfitting problems~\cite{rice2020overfitting}.

\paragraph{Learning in the Frequency Domain.} Frequency analysis provides a new perspective on the generalization behavior of DNNs. In~\cite{NIPS2015_536a76f9}, spectral pooling is designed to preserve more information than regular spatial-domain pooling. Tao \emph{et al.}~\cite{tao2022fat} \textcolor{black}{propose} a frequency-aware plug-in to remove redundant information effectively for quantization. Wang \emph{et al.}~\cite{wang2020high} {claim} CNNs could capture human-imperceptible high-frequency components of images for prediction, and smooth convolutional kernels are beneficial for robustness. 

\setlength{\tabcolsep}{0.9mm}
\section{Analysis}
\begin{table}
\footnotesize
\caption{Top-1 accuracy(\%)  of natural and robust ResNet18 models trained on CIFAR-10. The LPF row denotes the filter bandwidths applied to the inputs. The higher the value, the more information is retained (i.e. 32 means no filtering).}
\label{FrequencyFilterInput}
\centering
\scriptsize
\setlength{\tabcolsep}{0.9mm}{
\begin{tabular}{lc|ccccc}
\toprule
 \centering{model} & LPF & 32 & 28 & 24 & 20 & 16 \\
\midrule
\multirow{2}{*}{\centering {Natural}} & Clean & 94.56 & 92.92 & 90.75 & 80.7 & 50.72 \\
 & PGD-20 & 0.0 & 2.17 & 23.15 & 39.25 & 32.15  \\
\midrule
\multirow{2}{*}{\centering{Robust}} & Clean & 80.55 & 80.50 & 80.17 & 79.27 & 77.4 \\
 & PGD-20 & 51.81 & 52.10 & 52.22 & 52.54 & 52.52 \\ 
\bottomrule
\end{tabular}}
\end{table}
\paragraph{Reason for Low Standard Accuracy.} To explore the importance of high- and low-frequency information for models, we apply different LPFs to the natural and adversarial inputs that are fed into the natural or robust models to calculate the corresponding standard and robust accuracies. The results are shown in Table~\ref{FrequencyFilterInput}, e.g., a LPF bandwidth of 16 means after a Fast Fourier Transform (FFT), only the $16\times16$ patch in the center (viz. low frequencies) is preserved, and all external values are zeroed. As the bandwidth of LPF decreases, the information retained in the images also decreases. For the robust model, even though a large amount of high-frequency information is removed, there is only a negligible reduction in standard accuracy and less than a 1\% improvement in robust accuracy. This indicates that the robust model focuses primarily on low-frequency content for predictions, and the adversarial inputs rely on low-frequency components to exercise its aggressiveness. 
Furthermore, the standard accuracy of the robust model ($\approx$80\%) is similar to that of the natural model fed with natural inputs at LPF 20 (80.7\%). Such observation indicates that {\emph{the low standard accuracy in the robust model is due to the under-utilization of high-frequency components}}.

\paragraph{White-box Attack.}
In the natural model, as high-frequency information is removed, standard accuracy drops sharply. This suggests that the natural model employs high-frequency information to make classification judgment, which is consistent with the findings of~\cite{wang2020high}. Moreover, the accuracy subject to adversarial inputs is improved at lower LPF bandwidths, reflecting that the adversarial inputs of the natural model exhibit aggressiveness in both the high- and low-frequency regions. 

For robust models that focus on the low-frequency information, the aggressiveness of the adversarial inputs is mainly concentrated in the low-frequency region. Whereas for natural models that utilize both high- and low-frequency information, the hostility is embedded in both high- and low-frequency regions. This suggests that the white-box attack can adapt its aggressive frequency distribution to the target model’s frequency bias, thereby explaining why white-box attacks are so hard to defend.

\begin{figure}
\centering
\includegraphics[width=0.95\linewidth]{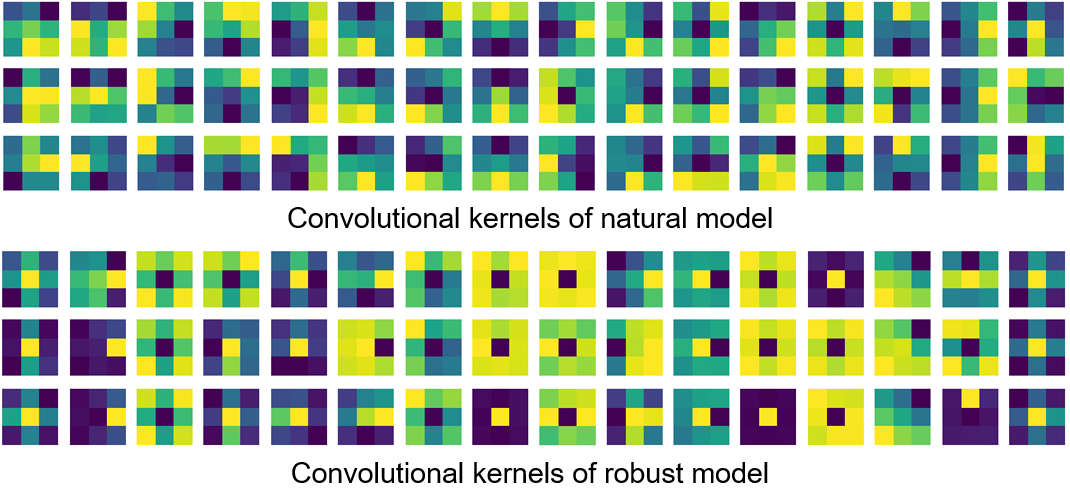}
\caption{Visualization of first convolutional kernels (16 kernels each channel $\times$ 3 channels) of a natural (top) and robust (bottom) model. The latter has smoother kernels than the former.} 
\label{fig:kernel}
\end{figure}
\paragraph{Smooth Kernels.} \textcolor{black}{Wang \emph{et al.}~\cite{wang2020high} introduce the concept of ``smooth" kernel which has a smooth envelope on its spatial weights. If a kernel is smooth, it will see a reduced amount of high-frequency information. Along this line, it is articulated that smoothing the kernels' adjacent spatial values can help improve the adversarial robustness.}
Since the kernels of the first layer deal directly with the images, they can respond to the frequency bias of the information extracted from the images. \textcolor{black}{In Figure~\ref{fig:kernel}, we visualize 16 randomly selected 3-channel kernels from the first layer of a natural (upper) and a robust model (lower), wherein the spatial size of each kernel is $3\times 3$.} 
The figure shows that the adjacent weights of kernels in the robust model
change less dramatically, producing smoother kernels than the natural model counterparts.
This implies that the robust model pays more attention to the low-frequency information, consistent with our previous discussion.


On the other hand, SWA~\cite{izmailov2018averaging}, which averages the values of weights over time (epochs) along the natural training trajectory, proves to be an effective method to improve the generalization of the models. Here, we utilize SWA in AT as a method of \emph{smoothing kernels} in the training time-axis dimension to mitigate the robust overfitting problems~\cite{chen2020robust}. Ablation study is shown in experiments to confirm the benefits of SWA in terms of robustness.

\section{Frequency Regularization}
\begin{figure}[t]
  \centering
  %
    \includegraphics[width=1\linewidth]{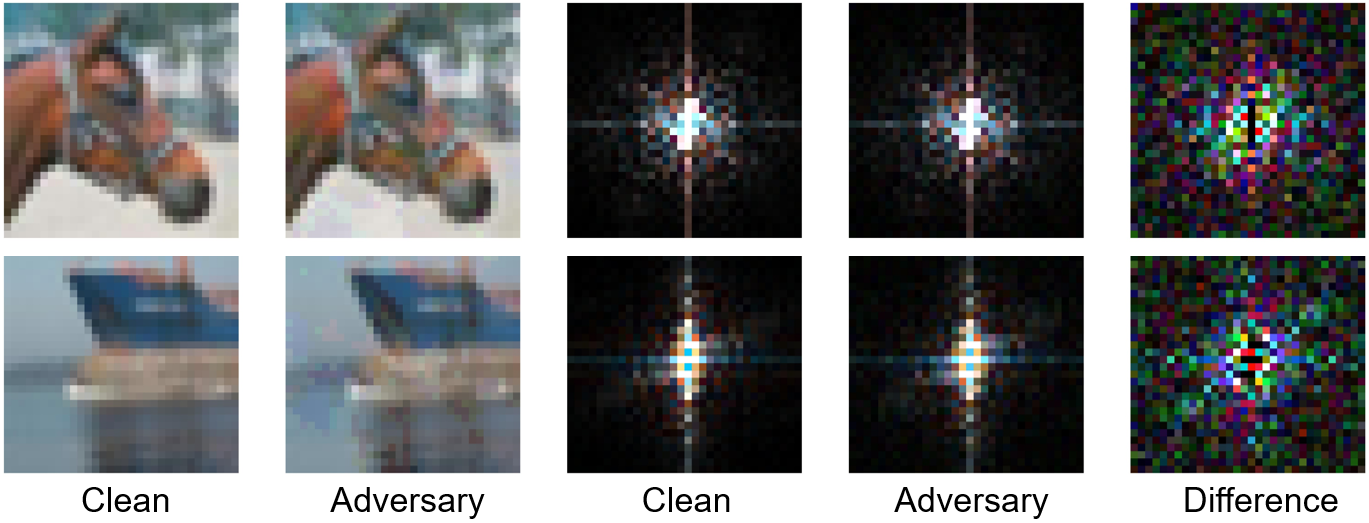}
   \caption{Visualization of natural and adversarial inputs in the spatial (left) and frequency domains (middle), and the absolute difference (right) after normalization in the frequency domain, with low frequency in the center and high frequency around. The brighter the pixel, the higher the frequency amplitude. The differences are mainly concentrated in the low-frequency region.}
   \label{AENE}
\end{figure}
To narrow the gap between standard and robust accuracies of the robust model, we need to identify the differences between natural and adversarial inputs. Figure~\ref{AENE} illustrates the natural and adversarial inputs in the spatial and spectral domains based on an adversarially trained model on the CIFAR-10 dataset. Because the adversarial inputs need to satisfy the $l_\infty$ norm constraints, the changes in the spatial domain are rather small, and one can still recognize the horse and the ship before and after the perturbation. In the frequency domain, as shown in Figure~\ref{AENE}, the differences between the natural and adversarial inputs are mainly distributed in the low-frequency region, with smaller amplitudes in the high-frequency region. Combined with the previous findings that robust models rely primarily on low-frequency information for prediction, it is easy to understand that the differences in the low-frequency region lead to a large accuracy gap. This further validates that, for the robust model, adversarial inputs rely mainly on low-frequency information to execute their aggressiveness. 

Inspired by these findings, we propose that if a model can be trained to limit such frequency differences and achieve similar spectral domain outputs, then robust accuracy can be improved by approaching standard accuracy.
To do so, we devise a simple yet effective frequency regularization (FR) to align the outputs for natural and adversarial inputs in the frequency domain. The optimization goal of the proposed AT with FR is:
\begin{equation}
    \mathcal{L}_{AT} = \mathcal{L}_{CE} + \lambda \cdot \frac{1}{n} \sum_{i=1}^{n}\emph{Dis}(\mathcal{F}(f(x_{i})), \mathcal{F}(f(x_{i}^{\prime}))),
\end{equation}
where $\mathcal{L}_{CE}$ denotes the cross-entropy loss, $\lambda$ (defaulted at $0.1$) denotes the FR coefficient, $\emph{Dis}$ denotes the distance function 
($\mathcal{L}_{1}$ is used), and $\mathcal{F}$ denotes the Discrete Fourier Transform (DFT). The distance function is applied to the real and imaginary parts of the complex numbers after the DFT, respectively, and the results are summed.
With FR, the robust accuracy against the PGD-20 attack on CIFAR-10 is substantially improved from 55.01\% to 59.49\%. 

\begin{figure}
\centering
\includegraphics[width=1\linewidth]{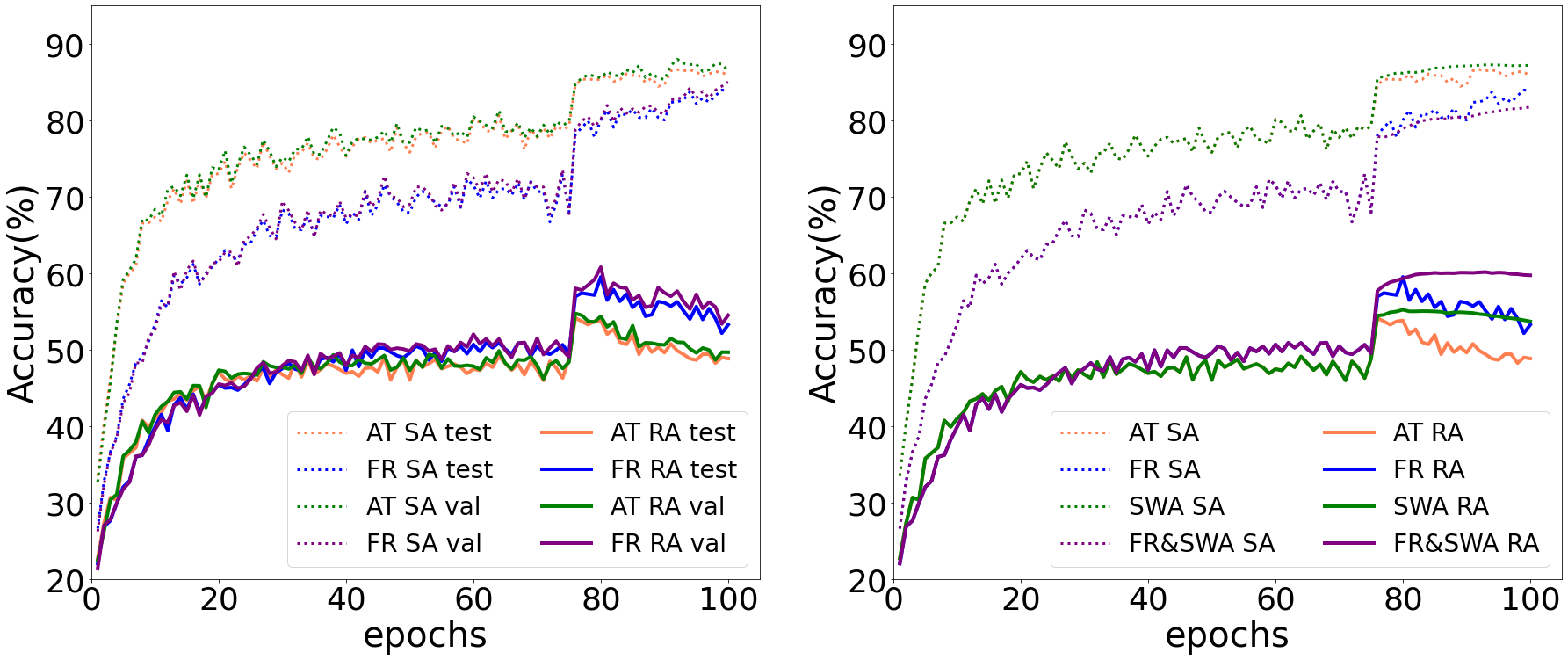}
\caption{\textbf{Left}: Standard accuracy (dashed line) and robust accuracy (solid line) on validation and test sets over epochs for AT-trained WideResNet-34-10 on CIFAR-10. FR denotes frequency regularization, SA and RA denote the standard and robust accuracies; \textbf{Right}: Ablation studies to demonstrate the effect of FR and SWA on model performance.}
\label{fig:ablation}
\end{figure}


\section{Experiments}
\paragraph{Experimental Settings.}
We take WideResNet-34-10 as a default model and adopt an SGD optimizer with a momentum of 0.9 and a global weight decay of $5\times10^{-4}$. The model is trained for 100 epochs with a batch size of 128 on one 3090 GPU. The initial learning rate is 0.1, decays to one-tenth at 75th and 90th epochs, respectively. All experiments are performed on the CIFAR-10 dataset, which contains 50k training (randomly split into a training set and a validation set at a 9:1 ratio) and 10k test examples. No extra data are used. We use PGD-10 AT as a standard training method. The robust accuracy of the PGD-20 attack equipped with random-start is taken as the main basis for robustness analysis. The attack step size is $\alpha$ = 2/255 and maximum $l_\infty$ norm-bounded perturbation $\epsilon$ = 8/255. SWA is used since the first epoch where the learning rate drops and continues until the end with a cycle length 1.

\paragraph{Experimental Results.}
We evaluate the robust accuracy against several popular attack methods, including FGSM~\cite{goodfellow2014explaining}, PGD~\cite{madry2017towards}, C\&W~\cite{carlini2017towards} and AA~\cite{croce2020reliable}, shown in Table~\ref{table:ablation}.
Following the default setting of AT, \textcolor{black}{the} 
attack step size is 2/255, and the maximum $l_\infty$ bounded perturbation is 8/255.
The standard and robust accuracies are used as the evaluation metrics.

As shown in Figure~\ref{fig:ablation}, our method succeeds in closing the gap from 29.61\% to 20.97\% with a 5.11\% improvement in robust accuracy against PGD-20 and a 3.53\% drop in standard accuracy. This matches the generally accepted theory that there is a trade-off between standard and robust accuracies. The ablation experiments show that FR (59.49\%) plays a major role in improving robust accuracy, while SWA (55.18\%) is utilized here to alleviate the overfitting problem. \textcolor{black}{The scheme that combines both of them} 
achieves the best 60.12\% and 54.35\% robust accuracy against PGD-20 attack and Autoattack, respectively.


\setlength{\tabcolsep}{0.9mm}
\begin{table}
\footnotesize
\caption{Top-1 robust accuracy(\%) of the WideResNet-34-10 model on the CIFAR-10. Bold numbers indicate the best.}

\label{table:ablation}
\centering
\scriptsize
\setlength{\tabcolsep}{0.9mm}{
\begin{tabular}{l|ccccc}
\toprule
 \centering{Method} & Clean & FGSM & PGD-20 & C\&W & AA  \\
\midrule
PGD-AT~\cite{rice2020overfitting} & 84.62 & 60.17 & 55.01 & 53.32 & 51.42  \\
TRADES~\cite{zhang2019theoretically} & 84.65 & 61.32 & 56.33 & 54.20 & 53.08  \\ 
MART~\cite{wang2019improving} & 84.17 & 61.61 & 58.56 & 54.58 & 51.10  \\ 
AWP~\cite{wu2020adversarial} & 85.57 & \textbf{62.90} & 58.14 & 55.96 & 54.04  \\
AT-SWA & \textbf{86.17} & 61.20 & 55.18 & 54.57 & 52.25 \\

\midrule

AT-FR(ours) & 80.59 & 61.47 & 59.49 & 54.33 & 52.06 \\
AT-FR-SWA(ours) & 81.09 & 62.49 & \textbf{60.12} & \textbf{56.14} & \textbf{54.35} \\

\bottomrule
\end{tabular}}
\end{table}

\section{Conclusion}
This work reveals that an adversarially trained model focuses primarily on low-frequency content for predictions, which accounts for the low standard accuracy due to under-utilization of high-frequency information. To this end, we devise a frequency regularization to align the logits for natural and adversarial inputs in the spectral domain. SWA is adopted temporally to smooth the weights,  improving the robustness further. Experiments show that the proposed method can substantially improve the robust accuracy. We further find that the white-box attack can adapt its aggressive frequency distribution to the target model’s frequency bias, which explains why white-box attacks are hard to defend. It is believed these findings can shed light on the design of  robust DNNs.

\section{Acknowledgement}
This work is supported in part by the Research Grants Council of the Hong Kong Special Administrative Region, China, under the General Research Fund (GRF) projects 17206020 and 17209721.

\bibliography{aaai23}

\end{document}